\documentclass[conference]{IEEEtran}
\IEEEoverridecommandlockouts

\usepackage{cite}
\usepackage{amsmath,amssymb,amsfonts}
\usepackage{graphicx}
\usepackage{subfigure}
\usepackage{wrapfig}
\usepackage{booktabs}
\usepackage{url}
\usepackage[hidelinks]{hyperref}

\begin{document}

\title{Auto-GlaKG: Automated Biomarker Reasoning for Explainable Glaucoma Detection%
\thanks{C. Huang is with the Department of Genetics, Washington University in St. Louis, St. Louis, MO 63110, USA (e-mail: chehua@wustl.edu).
Z. Han is with the Department of Computer Science, Southern Methodist University, Dallas, TX 75205, USA (e-mail: zeyuhan@smu.edu).
Y. Hu is with the Department of Health Data Science and Artificial Intelligence, University of Texas Health Science Center at Houston, Houston, TX 77030, USA (e-mail: yian.hu@uth.tmc.edu).
Y. Ding is with the Houston Methodist Neal Cancer Center, Houston Methodist Research Institute, Houston, TX 77030, USA (e-mail: yding2@houstonmethodist.org).
J. Qiu is with the Department of Data Science and Analytics, Georgetown University, Washington, DC 20057, USA (e-mail: jq167@georgetown.edu).
Y. Liu is with the Department of Banking and Finance, Nanyang Technological University, Singapore 639798 (e-mail: yadi001@ntu.edu.sg).}%
\thanks{J. Qiu and Y. Liu are the corresponding authors. Code: \url{https://github.com/CH-YellowOrange/Auto-GlaKG}.}%
}

\author{\IEEEauthorblockN{Cheng Huang, Zeyu Han, Yian Hu, Yaopeng Ding, Jingxi Qiu, and Yadi Liu}}

\maketitle

\begin{abstract}
Interpretable glaucoma detection requires a transparent connection between
fundus images, clinically meaningful biomarkers, and diagnostic predictions.
Knowledge graph (KG) reasoning provides this structure, but performance
measured using annotated biomarkers does not establish performance when
only images are available. We present Auto-GlaKG, an automated framework
that combines image-based biomarker prediction with fixed clinical-prior
rules. A ResNet50 network predicts the cup-to-disc ratio (CDR) and six
morphological signs using biomarker supervision, and a KG reasoner
aggregates their signed rule contributions into a glaucoma score. On 689
fundus images evaluated using stratified, group-aware five-fold
cross-validation, Auto-GlaKG achieves a pooled area under the receiver
operating characteristic curve (AUC) of 0.951 (95\% confidence interval:
0.931--0.969), with sensitivity of 0.970 and specificity of 0.779.
The matched image classifier also achieves an AUC of 0.951, while the
rule stage adds 0.2~ms per image. Replacing predicted CDR with its
annotation increases fold-mean AUC from 0.953 to 0.995. The framework
provides an explicit rule-level account of image-based predictions and
identifies CDR estimation as a priority for validation using
independently measured biomarkers.
\end{abstract}


\noindent\textit{Clinical Relevance}---Auto-GlaKG exposes predicted
optic-nerve findings and their contributions to an automated assessment,
enabling inspection of the evidence and of errors arising during
biomarker estimation.


\section{Introduction}

Glaucoma causes irreversible vision loss, motivating automated methods
for identifying suspicious optic-nerve changes in fundus
images~\cite{glaucoma2024global}. Deep learning has achieved promising
discrimination~\cite{glaucoma2025review,glaucoma2026xai}, but a diagnostic score
alone provides limited information about the anatomical findings driving
the prediction \cite{huang2026ai}. Linking image analysis to recognizable biomarkers,
including the cup-to-disc ratio (CDR), neuroretinal rim thinning, and
inferior--superior--nasal--temporal (ISNT) rule violation, can make the
computational decision process easier to
inspect~\cite{yang2025guidelines,glaucoma2026xai}.

Concept bottleneck models represent decisions through intermediate
concepts~\cite{liu2025hybridcbm,glaboost}, while knowledge graphs
(KGs) organize biomarkers and their diagnostic relationships. Our
previous GlaKG framework~\cite{glakg} explored structured
biomarker reasoning using annotations available at inference. That
setting establishes what the reasoner can do with supplied biomarker
information; it does not assess the complete image-based pipeline. This
distinction is especially important when biomarker descriptions are
AI-generated and closely aligned with diagnosis labels \cite{havasi2022leakage}.

We introduce Auto-GlaKG to evaluate automated biomarker reasoning under
image-only inference. The framework predicts seven biomarkers and
applies a fixed, graph-structured rule set to generate a glaucoma score
and explicit rule contributions. Our contributions are:
\begin{itemize}
  \item An automated image-to-biomarker-to-score pipeline that requires
  no biomarker annotations at inference.

  \item A controlled comparison of annotation-based and image-based
  reasoning under group-aware cross-validation.

  \item An analysis of biomarker substitution, rule activation, and
  explanation agreement that identifies the main limitations of the
  automated pipeline on the evaluated dataset.
\end{itemize}

\section{Methodology}

\begin{figure*}[t]
\centering
\includegraphics[width=1.0\textwidth]{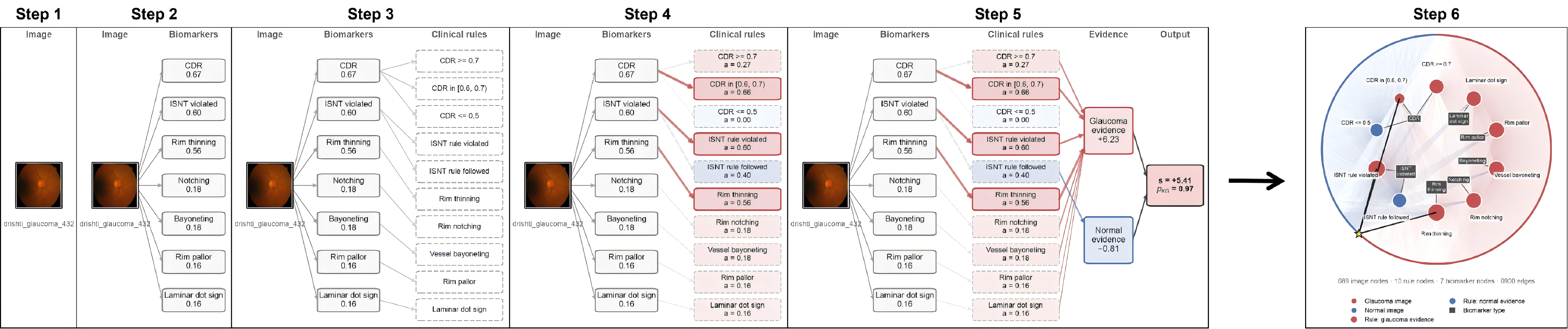}
\caption{Construction of an image-level evidence subgraph in Auto-GlaKG.
Step~1: the fundus image enters the graph as an image node.
Step~2: the biomarker network, trained without a diagnosis loss, predicts
seven biomarkers, the CDR and six morphological signs, which become
biomarker nodes.
Step~3: each biomarker is linked to the clinical rules defined on it
through schema-level relations, yielding ten candidate rules.
Step~4: rule activations $a_r$ are computed softly from the predicted
biomarkers (red: glaucoma evidence; blue: normal evidence; dashed:
$a_r<0.5$).
Step~5: activated rules contribute strength-weighted evidence to the
glaucoma-directed and normal-directed sums, whose signed difference
defines the KG score $s$ and the probability $p^{\mathrm{KG}}$.
Step~6: the subgraph joins the global KG through the shared rule and
biomarker nodes (the example is marked by a star). Highlighted rules and
edges provide a compact visualization; all soft contributions, including
those below the visualization threshold, remain in the score.}
\label{fig:construction}
\end{figure*}

\subsection{Biomarker-Supervised Image Network}

Let $x_i$ denote a fundus image, $y_i\in\{0,1\}$ its diagnosis with
glaucoma as the positive class, and
$\mathbf{b}_i=(c_i,b_{i1},\ldots,b_{i6})$ its supplied biomarker targets.
The targets comprise CDR $c_i\in[0,1]$, ISNT violation, rim thinning,
notching, vessel bayoneting, rim pallor, and the laminar dot sign.
An ImageNet-pretrained ResNet50~\cite{resnet} with a seven-output head
predicts $\hat{\mathbf{b}}_i=f_\theta(x_i)$, consisting of CDR
$\hat{c}_i$ and six sign probabilities $\hat{p}_{ik}$.

The network is optimized using
\begin{equation}
\mathcal{L}_{\mathrm{bm}}
=5\,\ell_{\mathrm{SL1}}(\hat{c}_i,c_i)
+\frac{1}{6}\sum_{k=1}^{6}
\ell_{\mathrm{BCE}}(\hat{p}_{ik},b_{ik};\pi_k),
\label{eq:loss}
\end{equation}
where $\ell_{\mathrm{SL1}}$ is smooth L1 loss and
$\ell_{\mathrm{BCE}}$ is weighted binary cross-entropy. Positive-class
weights $\pi_k$ are derived from the training-fold biomarker prevalence;
missing targets are masked. The CDR weight of 5 balances the smaller
scale of the regression loss against the classification terms.
This network uses biomarker supervision without an explicit
diagnosis-classification loss.

\subsection{Graph-Structured Clinical Rules}

\begin{wraptable}{r}{0.52\columnwidth}
\vspace{-12pt}
\centering
\tiny
\caption{Fixed rule set. G and N indicate glaucoma-directed
($e_r=+1$) and normal-directed ($e_r=-1$) evidence.}
\label{tab:rules}
\setlength{\tabcolsep}{2.5pt}
\begin{tabular}{llcc}
\hline
Rule & Condition & Dir. & $w_r$ \\
\hline
CDR high & $c\ge0.7$ & G & 3 \\
CDR moderate & $0.6\le c<0.7$ & G & 2 \\
CDR normal & $c\le0.5$ & N & 3 \\
Rim thinning & Present & G & 3 \\
Notching & Present & G & 3 \\
ISNT violated & Present & G & 2 \\
ISNT followed & No violation & N & 2 \\
Bayoneting & Present & G & 2 \\
Rim pallor & Present & G & 1 \\
Laminar dot sign & Present & G & 1 \\
\hline
\end{tabular}
\vspace{-10pt}
\end{wraptable}

The KG contains image nodes, seven biomarker types, and ten rule nodes.
Biomarker-to-rule relations encode the rule definitions, while weighted
image-to-rule edges represent activation strengths
(Fig.~\ref{fig:construction}). The graph organizes the evidence;
prediction is computed by fixed additive rule aggregation.

Each rule has a condition $\mathcal{C}_r$, an evidence direction
$e_r\in\{+1,-1\}$, and a strength $w_r\in\{3,2,1\}$
(Table~\ref{tab:rules}). These settings encode simplified clinical priors
and are fixed for the experiments. A CDR in $(0.5, 0.6)$ activates no
CDR rule under hard activation and is treated as indeterminate.

\subsection{Soft Activation and Rule Contributions}

Hard activation applies the rule conditions directly. To reduce
discontinuities near CDR thresholds, soft activation uses
\begin{equation}
a_{ir}=
\begin{cases}
\sigma((\hat{c}_i-\tau_r)/T), & r=\text{CDR high},\\
\sigma((\tau_r-\hat{c}_i)/T), & r=\text{CDR normal},\\
\hat{p}_{ik}, & r \text{ requires sign } k \text{ present},\\
1-\hat{p}_{ik}, & r \text{ requires sign } k \text{ absent},
\end{cases}
\label{eq:activation}
\end{equation}
where $\tau_r$ is the threshold in Table~\ref{tab:rules} (0.7 and 0.5),
$\sigma$ is the logistic function, and $T=0.03$. The moderate-CDR
rule uses
\begin{equation}
a_{i,\mathrm{mod}}=
\sigma\!\left(\frac{\hat{c}_i-0.6}{T}\right)
\sigma\!\left(\frac{0.7-\hat{c}_i}{T}\right).
\end{equation}
Away from the CDR boundaries, these sigmoid memberships approach their
hard-threshold counterparts as $T\to0$. Binary-sign rules use their
predicted probabilities directly. $T$ is fixed for all experiments, and
its effect is reported in Table~\ref{tab:ablation}(a).

The signed contribution of rule $r$ and the resulting prediction are
\begin{equation}
\begin{aligned}
s_i&=\sum_r\phi_{ir},\qquad \phi_{ir}=e_rw_ra_{ir},\\
p_i^{\mathrm{KG}}&=\sigma(s_i/\lambda),
\end{aligned}
\label{eq:score}
\end{equation}
with $\lambda=1.5$ and a positive decision when
$p_i^{\mathrm{KG}}\ge0.5$. Because this decision is equivalent to
$s_i\ge0$, $\lambda$ affects calibration only, not AUC or
threshold-dependent metrics. The reasoner has no trainable parameters,
and its logistic output is evaluated as a prediction score.

The complete contribution vector exactly reconstructs $s_i$.
For visualization, rules with $a_{ir}\ge0.5$ are highlighted and ordered
by $|\phi_{ir}|$; all remaining contributions are retained in scoring.

\subsection{Image Fusion and Analysis Settings}

An optional fusion combines the KG score with a separate
diagnosis-supervised ResNet50:
\begin{equation}
p_i^{\mathrm{fus}}=
\alpha p_i^{\mathrm{KG}}+(1-\alpha)p_i^{\mathrm{img}}.
\end{equation}
The weight $\alpha\in\{0,0.1,\ldots,1\}$ is selected by inner-validation
AUC. Rule contributions describe the KG branch; they do not fully
explain the image classifier or the fused prediction.

\begin{table*}[t]
\centering
\caption{Diagnostic Performance on Pooled Out-of-Fold Predictions. AUC
Intervals Are 95\% Group-Bootstrap Confidence Intervals; Fold AUC Is
Mean $\pm$ Std over the Five Test Folds. Best Value per Column in Bold
(Lower Is Better for Brier and ECE). Acc: Accuracy; BalAcc: Balanced
Accuracy; Sens: Sensitivity; Spec: Specificity; Prec: Precision; MCC:
Matthews Correlation Coefficient; ECE: Expected Calibration Error;
Sens@95Spec: Sensitivity at 95\% Specificity.}
\label{tab:main}
\setlength{\tabcolsep}{3.5pt}
\resizebox{\textwidth}{!}{%
\begin{tabular}{l|ccccccccccccc}
\hline
Method & AUC [95\% CI] & Fold AUC & AUPRC & Acc & BalAcc & Sens & Spec & Prec & F1 & MCC & Brier$\downarrow$ & ECE$\downarrow$ & Sens@95Spec \\
\hline
ResNet50
& \textbf{0.951} [0.929, 0.970] & $0.952\pm0.028$ & 0.955 & 0.910 & \textbf{0.907} & 0.920 & \textbf{0.893} & \textbf{0.934} & 0.927 & 0.810 & 0.075 & 0.058 & \textbf{0.724} \\
\hline
Auto-GlaKG (hard)
& 0.938 [0.914, 0.960] & $0.938\pm0.025$ & 0.943 & 0.900 & 0.888 & 0.937 & 0.840 & 0.905 & 0.921 & 0.786 & 0.088 & 0.075 & 0.604 \\
Auto-GlaKG (soft)
& \textbf{0.951} [0.931, 0.969] & $0.953\pm0.023$ & \textbf{0.962} & 0.897 & 0.874 & \textbf{0.970} & 0.779 & 0.877 & 0.921 & 0.782 & 0.087 & 0.090 & 0.658 \\
\hline
Auto-GlaKG (hard) + Image
& 0.949 [0.927, 0.968] & $0.952\pm0.026$ & 0.961 & 0.903 & 0.897 & 0.920 & 0.874 & 0.923 & 0.922 & 0.794 & 0.082 & 0.070 & 0.700 \\
Auto-GlaKG (soft) + Image
& 0.946 [0.925, 0.967] & $\mathbf{0.955\pm0.025}$ & 0.955 & \textbf{0.912} & 0.901 & 0.944 & 0.859 & 0.916 & \textbf{0.930} & \textbf{0.811} & \textbf{0.072} & \textbf{0.044} & 0.600 \\
\hline
\end{tabular}}
\end{table*}

The annotation-based, or \emph{oracle}, setting applies the same
current reasoner to $\mathbf{b}_i$ instead of
$\hat{\mathbf{b}}_i$. Here the oracle denotes access to the supplied
annotations, not independently verified clinical measurements.
Single-biomarker substitution replaces one prediction with its
annotation to measure the corresponding change in AUC. Hard-rule flips
compare rule conditions evaluated on predicted and annotated biomarkers.
Explanation agreement is the Jaccard index of the two sets of
highlighted rules:
\begin{equation}
J_i=
\frac{|\mathcal{R}^{\mathrm{ann}}_i\cap
\mathcal{R}^{\mathrm{pred}}_i|}
{|\mathcal{R}^{\mathrm{ann}}_i\cup
\mathcal{R}^{\mathrm{pred}}_i|},
\end{equation}
with $J_i=1$ when both sets are empty.
These annotation-dependent analyses are retrospective diagnostics.

\section{Experimental Setup}

\subsection{Dataset and Evaluation Protocol}
\subsubsection{Dataset}
We use a public dataset\footnote{\url{https://huggingface.co/datasets/AswanthCManoj/glaucoma_diagnosis_json_analysis}}
containing 689 fundus images: 427 glaucoma and 262 normal images drawn
from ACRIMA~\cite{diaz2019cnns}, RIM-ONE~\cite{fumero2011rimone}, and
Drishti-GS~\cite{sivaswamy2014drishti}. Seven biomarker targets are parsed
from the accompanying AI-generated descriptions. Risk categories and
annotation confidence scores are not inputs to the automated reasoner.

\subsubsection{Evaluation Protocol}
Perceptual-hash grouping with Hamming distance $\le4$ yields 652 groups,
including 37 two-image groups. Stratified, group-aware five-fold
cross-validation keeps detected near-duplicates in the same fold.
Within each outer training fold, a group-aware split reserves 20\% for
fusion-weight selection, and both networks are trained on the remaining
80\%. Outer test folds are excluded from training and fusion selection.

\subsection{Training and Evaluation}
\subsubsection{Training}
Images are resized to $224\times224$ and augmented using rotations
($\pm10^\circ$), scaling (0.9--1.1), translation ($\pm6\%$), horizontal
flipping, and brightness and contrast jitter. Vertical flipping is
excluded to preserve superior--inferior orientation. Both networks use
AdamW with learning rate $10^{-4}$, weight decay $10^{-4}$, batch size
32, cosine learning-rate decay, and 15 training epochs.
Experiments use PyTorch, seed 42, and four NVIDIA A100 GPUs (80~GB).
Each ResNet50 has approximately 23.5 million parameters; reported
inference times are 6.3~ms for one network and 0.2~ms for rule evaluation.
Fusion additionally requires the separate image classifier.

\subsubsection{Evaluation}
We compare the image classifier, hard and soft Auto-GlaKG,
and image fusion. Threshold-dependent metrics use a cutoff of 0.5, and AUC
confidence intervals are obtained from 1,000 group-level bootstrap
resamples. Fold-mean AUC is distinguished from pooled AUC throughout.
Calibration is assessed by expected calibration error (ECE; ten bins).
Exploratory AUC comparisons use paired DeLong
tests~\cite{delong1988comparing} with Holm correction.
Biomarker agreement is measured using CDR mean absolute error (MAE),
Pearson correlation, and Cohen's $\kappa$ for binary signs.

\section{Results and Discussion}

\subsection{Automated and Annotation-Based Performance}

Table~\ref{tab:main} reports pooled out-of-fold
metrics. With image-only inference, soft Auto-GlaKG achieves a pooled AUC of
0.951 (fold mean $0.953\pm0.023$), matching the image classifier
(0.951). The DeLong test detects no difference ($\Delta=-0.0001$,
$p=0.99$), and the 95\% confidence interval of the difference,
$[-0.020, 0.020]$, excludes changes larger than 0.02 in either
direction. This bounds the difference but does not formally establish
equivalence. At the default threshold, Auto-GlaKG favors sensitivity
(0.970 versus 0.920; 13 versus 34 missed glaucoma cases) at the cost of
specificity (0.779 versus 0.893; 58 versus 28 false positives), a
trade-off that suits screening but increases the referral burden.
This advantage does not extend to high-specificity operation: at 95\%
specificity, soft Auto-GlaKG reaches a sensitivity of 0.658 versus 0.724
for the image classifier, and its ECE (0.090) is the highest among all
settings.

Soft fusion yields an AUC of 0.946, a marginally higher MCC (0.811
versus 0.810), and the lowest ECE (0.044) among all settings, but it
does not improve discrimination. The selected fusion weight also varies
widely across folds ($\alpha=0.42\pm0.49$), so no stable fusion
preference emerges under this protocol.

\begin{figure}[ht]
\centering

\subfigure[Annotated Biomarkers]{
    \includegraphics[width=0.45\linewidth]{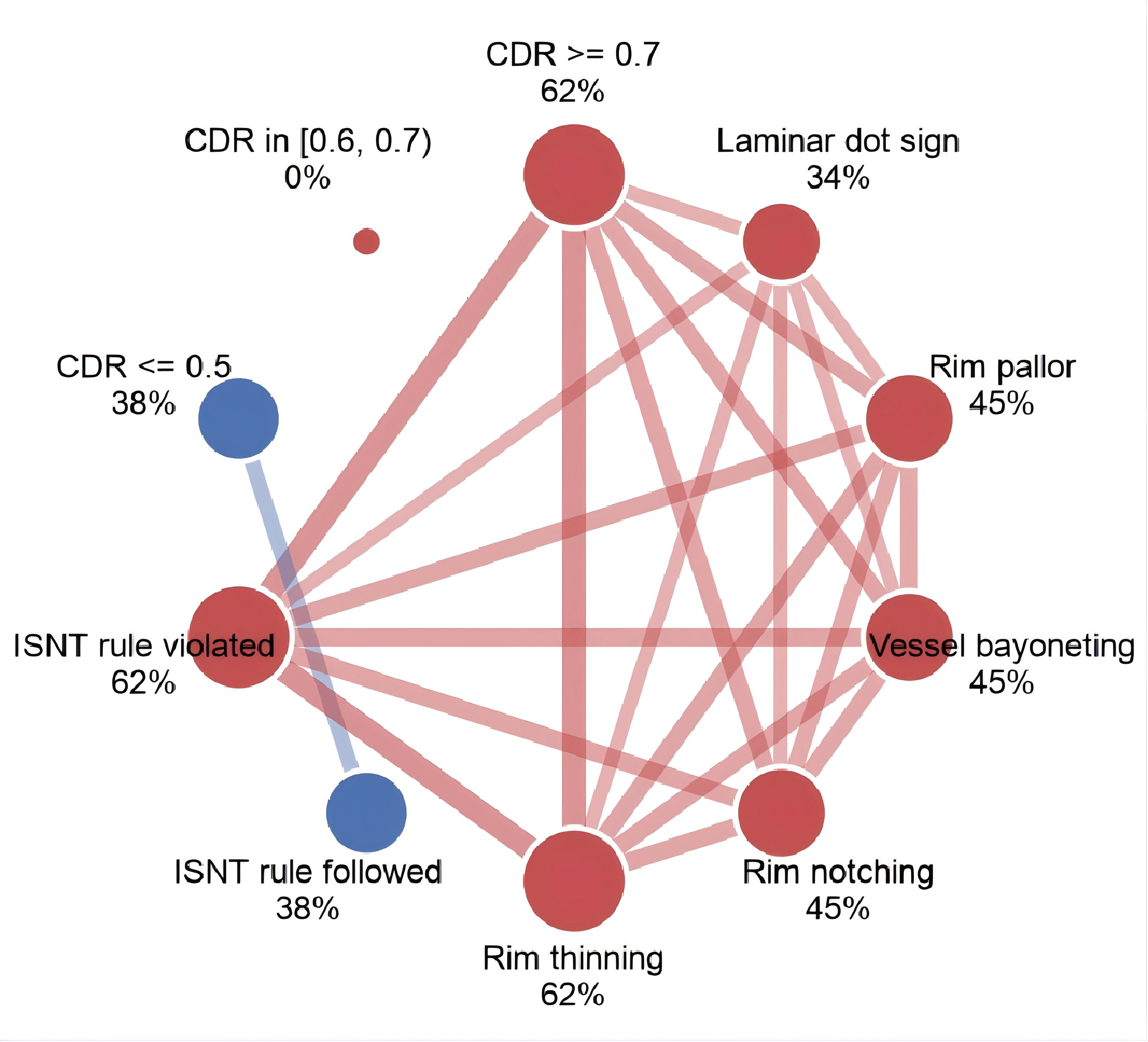}
    \label{fig:coact_a}
}
\hfill
\subfigure[Predicted Biomarkers]{
    \includegraphics[width=0.45\linewidth]{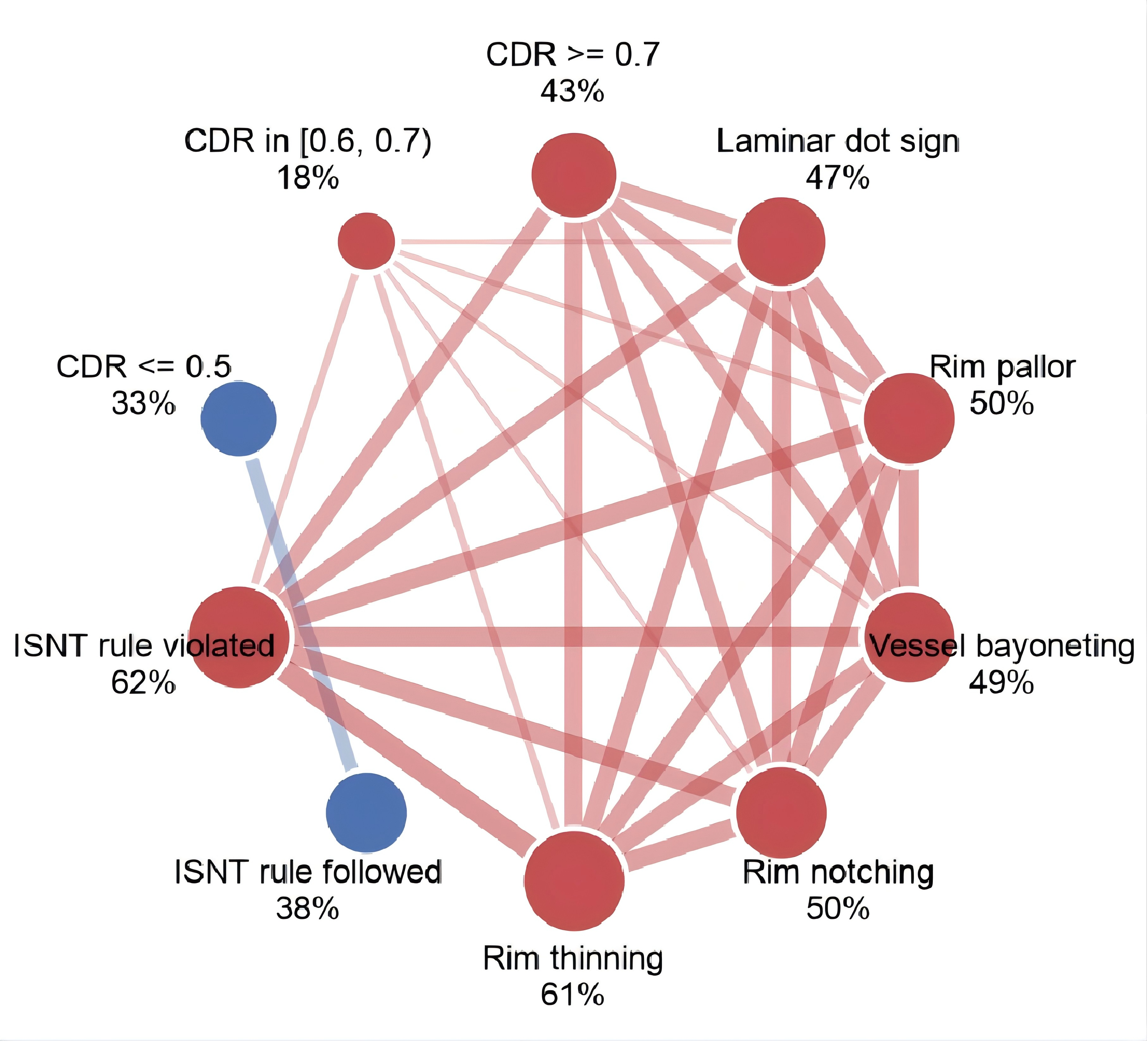}
    \label{fig:coact_b}
}

\caption{Rule co-activation networks. A rule is counted as activated when
$a_{ir}\ge0.5$. Node size is the activation frequency of a rule; edge
width is the fraction of images in which two rules are activated together
(red: both glaucoma evidence; blue: both normal evidence; grey: mixed).
Annotated biomarkers yield two disjoint, near-deterministic rule
families, whereas predicted biomarkers yield weaker and more
heterogeneous co-activations.}
\label{fig:coact}
\end{figure}

\begin{table*}[ht]
\centering
\caption{KG Ablations for Hard and Soft Activation. (a) Pooled
Out-of-Fold AUC under Different Rule Weights, Rule Subsets, Soft-Threshold
Temperatures, and Image Fusion. (b) Change in Pooled AUC when a Single
Rule Is Removed, and Fold-Mean AUC when a Single Predicted Biomarker Is
Replaced by Its Annotation; Percentages Give the Fraction of the Gap
between No Substitution and Annotation-Based Reasoning Recovered by That
Biomarker. Fold-Mean Substitution Results Must Be Compared with the
Fold-Mean Baseline (No Substitution). Defaults Are Marked with $^{\ast}$.}
\label{tab:ablation}
\footnotesize
\setlength{\tabcolsep}{4pt}
\begin{minipage}[t]{0.44\textwidth}
\centering
\textbf{(a)}\\[2pt]
\begin{tabular}{lcc}
\hline
Setting & Hard & Soft \\
\hline
\multicolumn{3}{l}{\textit{Rule weighting}} \\
Clinical 3/2/1$^{\ast}$ & 0.938 & 0.951 \\
Uniform 1/1/1 & 0.933 & 0.949 \\
Learned (logistic, cross-fold) & 0.925 & 0.947 \\
\hline
\multicolumn{3}{l}{\textit{Rule subset (number of rules)}} \\
CDR only (3) & 0.935 & 0.953 \\
+ Rim thinning (4) & 0.936 & 0.953 \\
+ ISNT (6) & 0.936 & 0.953 \\
+ Notching (7) & 0.937 & 0.952 \\
All seven biomarkers (10)$^{\ast}$ & 0.938 & 0.951 \\
\hline
\multicolumn{3}{l}{\textit{Image fusion}} \\
Without image$^{\ast}$ & 0.938 & 0.951 \\
With image & 0.949 & 0.946 \\
\hline
Temperature $T$ (soft) & AUC & ECE \\
\hline
0.005 & 0.951 & 0.082 \\
0.01 & 0.951 & 0.085 \\
0.03$^{\ast}$ & 0.951 & 0.090 \\
0.05 & 0.951 & 0.098 \\
0.08 & 0.951 & 0.106 \\
0.12 & 0.951 & 0.121 \\
\hline
\end{tabular}
\end{minipage}\hfill
\begin{minipage}[t]{0.54\textwidth}
\centering
\textbf{(b)}\\[2pt]
\begin{tabular}{lcc}
\hline
Setting & Hard & Soft \\
\hline
\multicolumn{3}{l}{\textit{Leave-one-rule-out ($\Delta$AUC, pooled)}} \\
CDR high & $-0.0192$ & $-0.0050$ \\
CDR moderate & $-0.0003$ & $+0.0009$ \\
CDR normal & $-0.0108$ & $-0.0009$ \\
Rim thinning & $+0.0001$ & $-0.0001$ \\
Notching & $-0.0002$ & $+0.0012$ \\
ISNT violated & $+0.0001$ & $-0.0001$ \\
ISNT followed & $+0.0001$ & $-0.0001$ \\
Bayoneting & $-0.0011$ & $+0.0006$ \\
Rim pallor & $-0.0002$ & $+0.0003$ \\
Laminar dot sign & $-0.0001$ & $+0.0005$ \\
\hline
\multicolumn{3}{l}{\textit{Annotation substitution (fold-mean AUC, gap recovered)}} \\
No substitution & $0.938\pm0.025$ & $0.953\pm0.023$ \\
Annotated CDR & $0.980\pm0.012$ (67\%) & $0.995\pm0.004$ (89\%) \\
Annotated ISNT violation & $0.977\pm0.013$ (63\%) & $0.986\pm0.009$ (71\%) \\
Annotated rim thinning & $0.974\pm0.014$ (57\%) & $0.980\pm0.011$ (58\%) \\
Annotated notching & $0.963\pm0.015$ (40\%) & $0.970\pm0.014$ (37\%) \\
Annotated bayoneting & $0.961\pm0.014$ (38\%) & $0.967\pm0.014$ (30\%) \\
Annotated rim pallor & $0.959\pm0.016$ (34\%) & $0.961\pm0.017$ (17\%) \\
Annotated laminar dot sign & $0.954\pm0.020$ (25\%) & $0.959\pm0.019$ (13\%) \\
All annotated (oracle) & $1.000\pm0.000$ & $1.000\pm0.000$ \\
\hline
\end{tabular}
\end{minipage}
\end{table*}

At the rule level, the co-activation networks in Fig.~\ref{fig:coact}
contrast annotated and predicted biomarkers: annotated biomarkers
activate two disjoint, near-deterministic rule families, whereas
predicted biomarkers produce weaker and more heterogeneous
co-activations.

\subsection{Biomarker Prediction and Ablation}

The predicted CDR has an MAE of 0.080 and a Pearson correlation of 0.81;
Cohen's $\kappa$ ranges from 0.50 (laminar dot sign) to 0.77 (ISNT
violation). Soft activation outperforms hard activation in every rule
configuration (Table~\ref{tab:ablation}(a); 0.951 versus 0.938 by
default), and $T$ affects calibration (ECE 0.082 to 0.121) rather than
AUC. The CDR rules alone already reach 0.953, and no single rule removal
changes the soft-activation AUC by more than 0.005, so the additional
signs enrich the explanation more than the discrimination. Fusion helps
only under hard activation (0.938 to 0.949). Annotation-based reasoning
reaches a fold-mean AUC of 1.000, which defines the upper end of the gap
in Table~\ref{tab:ablation}(b). Substituting the annotated CDR raises
the fold-mean AUC from 0.953 to 0.995, recovering about 89\% of the gap,
followed by ISNT violation (71\%) and rim thinning (58\%). These
fractions overlap and are not a causal decomposition, but they identify
the CDR as the main bottleneck.

\subsection{Rule Agreement \& Error Analysis}

For soft Auto-GlaKG + Image, flips of CDR, ISNT, and rim-thinning rules
are associated with more errors than flips of weaker rules.
Mean Jaccard agreement between predicted and annotation-derived rule sets
is 0.76: 0.82 for the 628 correctly classified images and 0.13 for the
61 errors. This association is descriptive, since agreement concerns the
KG branch and the outcome concerns the fused model, and it requires
reference annotations, so it cannot serve as a deployment-time error
detector.

\section{Conclusion}

Auto-GlaKG combines automated biomarker prediction with fixed,
graph-structured clinical rules for explainable glaucoma detection.
It achieves pooled AUC 0.951 on the evaluated dataset and provides an
exact decomposition of the KG score. Comparing annotation-based and
automated reasoning exposes the effects of biomarker prediction errors,
with CDR substitution producing the largest improvement in this dataset.
These findings motivate further evaluation with expert biomarker
measurements and independent data sources.

\section{Limitations \& Future Work}

\subsection{Limitations}

The AI-generated biomarker targets are strongly associated with the
diagnosis and have not been validated as clinical measurements, so
omitting a diagnosis loss does not guarantee label-independent learning.
The small, pooled dataset and hash-based grouping limit conclusions
about patient-independent and cross-source performance. Paired tests do
not fully account for within-group dependence, and inference times are
specific to our hardware and implementation.

\subsection{Future Work}

Future work will evaluate CDR derived from optic disc and cup
segmentation~\cite{fu2018joint,orlando2020refuge}, using expert
annotations and documented grading procedures. Independent-source
evaluation and ophthalmologist reader studies are needed to assess
generalization and the usefulness of the rule-level evidence.


\bibliographystyle{IEEEtran}
\bibliography{main}

\end{document}